\def\year{2021}\relax
\documentclass[letterpaper]{article} 
\usepackage{aaai21}  
\usepackage{times}  
\usepackage{helvet} 
\usepackage{courier}  
\usepackage[hyphens]{url}  
\usepackage{graphicx} 

\usepackage{amsmath}
\usepackage{multirow} 
\usepackage{multicol}
\usepackage{subfigure}
\usepackage{color}
\usepackage{amsfonts,amssymb} 
\usepackage{booktabs}
\usepackage[switch]{lineno}
\usepackage{natbib}  
\usepackage{caption} 
\title{Retrospective Reader for Machine Reading Comprehension}
\author{
   Zhuosheng Zhang\textsuperscript{\rm 1,2,3},
	Junjie Yang\textsuperscript{\rm 2,3,4},
	Hai Zhao\textsuperscript{\rm 1,2,3,\thanks{Corresponding author.  This paper was partially supported by National Key Research and Development Program of China (No. 2017YFB0304100), Key Projects of National Natural Science Foundation of China (U1836222 and 61733011), Huawei-SJTU long term AI project, Cutting-edge Machine reading comprehension and language model.}}\\
}
\affiliations{
    \textsuperscript{\rm 1}Department of Computer Science and Engineering, Shanghai Jiao Tong University\\
	\textsuperscript{\rm 2}Key Laboratory of Shanghai Education Commission for Intelligent Interaction\\
	and Cognitive Engineering, Shanghai Jiao Tong University, Shanghai, China\\
	\textsuperscript{\rm 3}MoE Key Lab of Artificial Intelligence, AI Institute, Shanghai Jiao Tong University, Shanghai, China\\
	\textsuperscript{\rm 4}SJTU-ParisTech Elite Institute of Technology, Shanghai Jiao Tong University, Shanghai, China\\
	{\tt
	zhangzs@sjtu.edu.cn, jj-yang@sjtu.edu.cn, zhaohai@cs.sjtu.edu.cn}
}

\begin{document}
\maketitle

\begin{abstract}
Machine reading comprehension (MRC) is an AI challenge that requires machines to determine the correct answers to questions based on a given passage. MRC systems must not only answer questions when necessary but also tactfully abstain from answering when no answer is available according to the given passage. When unanswerable questions are involved in the MRC task, an essential verification module called verifier is especially required in addition to the encoder, though the latest practice on MRC modeling still mostly benefits from adopting well pre-trained language models as the encoder block by only focusing on the ``reading''. This paper devotes itself to exploring better verifier design for the MRC task with unanswerable questions. Inspired by how humans solve reading comprehension questions, we proposed a retrospective reader (Retro-Reader) that integrates two stages of reading and verification strategies: 1) sketchy reading that briefly investigates the overall interactions of passage and question, and yields an initial judgment; 2) intensive reading that verifies the answer and gives the final prediction. The proposed reader is evaluated on two benchmark MRC challenge datasets SQuAD2.0 and NewsQA, achieving new state-of-the-art results. Significance tests show that our model is significantly better than strong baselines. 
\end{abstract}

\section{Introduction}
\begin{center}
	\begin{tabular}{p{7.8cm}}
		\noindent\emph{Be certain of what you know and be aware what you don't. That is wisdom.} \\
		\midrule
		\hfill{\emph{ Confucius {\rm(551 BC - 479 BC)}}}\\
	\end{tabular}
\end{center}

\begin{table}[htb]
	\begin{center}
		\begin{tabular}{|p{8.0cm}|}
			\hline
			\textbf{Passage}: \\ 
			\textit{Computational complexity theory is a branch of the theory of computation in theoretical computer science that focuses on classifying computational problems according to their inherent difficulty, and relating those classes to each other. A computational problem is understood to be a task that is in principle amenable to being solved by a computer, which is equivalent to stating that the problem may be \textcolor[RGB]{0,145,147}{solved by mechanical application of mathematical steps, such as an algorithm}}.\\ 
			\hline
			\textbf{Question}: \\
			\textit{What cannot be solved by mechanical application of mathematical steps}?\\    
			\hline
			\textbf{Gold Answer:} $\langle\textup{no answer}\rangle$ \\
			\textbf{Plausible answer:} \textit{algorithm} \\
			\hline
		\end{tabular}
	\end{center}
	\caption{\label{table01} An unanswerable MRC example.}
\end{table}

\begin{figure*}[htb]
	\centering
	\includegraphics[width=1.0\textwidth]{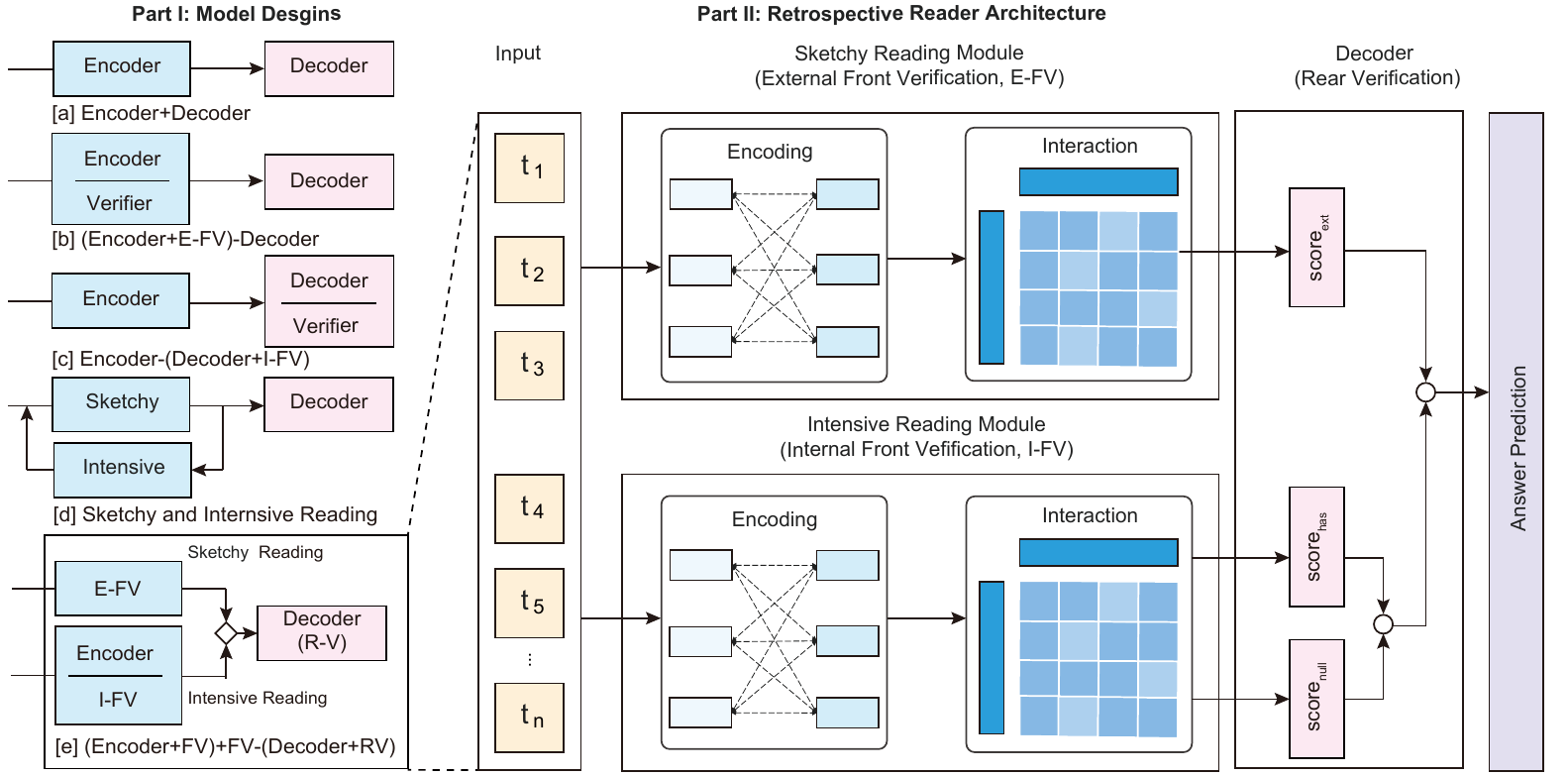}
	\caption{\label{fig:overview}  Reader overview. For the left part, models [a-c] summarize the instances in previous work, and model [d] is ours, with the implemented version [e]. In the names of models [a-e], ``($\cdot$)'' represents a module, ``+'' means the parallel module and ``-'' is the pipeline. The right part is the detailed architecture of our proposed Retro-Reader. }
\end{figure*}

Machine reading comprehension (MRC) aims to teach machines to answer questions after comprehending given passages \cite{hermann2015teaching,Joshi2017TriviaQA,Rajpurkar2018Know}, which is a fundamental and long-standing goal of natural language understanding (NLU) \cite{zhang2020mrc}. It has significant application scenarios 
such as question answering and dialog systems \cite{zhang2018modeling,choi2018quac,reddy2019coqa,zhang2018mrc,xu2020topic,zhu2018lingke}.
The early MRC systems \cite{kadlec2016text,Chen2016A,Dhingra2017Gated,Wang2017Gated,Seo2016Bidirectional} were designed on a latent hypothesis that all questions can be answered according to the given passage (Figure \ref{fig:overview}-[a]), which is not always true for real-world cases. 
The recent progress on the MRC task has required that the model must be capable of distinguishing those unanswerable questions to avoid giving plausible answers \cite{Rajpurkar2018Know}. MRC task with unanswerable questions may be usually decomposed into two subtasks: 1) answerability verification and 2) reading comprehension.
To determine unanswerable questions requires a deep understanding of the given text and requires more robust MRC models, making MRC much closer to real-world applications. Table \ref{table01} shows an unanswerable example from SQuAD2.0 MRC task \cite{Rajpurkar2018Know}. 


So far, a common reading system (reader) which solves MRC problem generally consists of two modules or building steps as shown in Figure \ref{fig:overview}-[a]:
1) building a robust language model (LM) as encoder; 2) designing ingenious mechanisms as decoder according to MRC task characteristics.

Pre-trained language models (PrLMs) such as BERT \cite{devlin2018bert} and XLNet \cite{yang2019xlnet} have achieved success on various natural language processing tasks, which broadly play the role of a powerful encoder \cite{zhang2019explicit,li2020explicit,zhou2019limit}. 
However, it is quite time-consuming and resource-demanding to impart massive amounts of general knowledge from external corpora into a deep language model via pre-training. 

Recently, most MRC readers keep the primary focus on the encoder side, i.e., the deep PrLMs \cite{devlin2018bert,yang2019xlnet,Lan2020ALBERT}, as readers may simply and straightforwardly benefit from a strong enough encoder. Meanwhile, little attention is paid to the decoder side\footnote{We define \textit{decoder} here as the task-specific part in an MRC system, such as passage and question interaction and answer verification.} of MRC models 
\cite{hu2019read,back2020neurquri,reddy2020answer}, though it has been shown that better decoder or better manner of using encoder still has a significant impact on MRC performance, no matter how strong the encoder it is \cite{dcmn20,liu2021filling,li2019dependency,li2018unified,zhu2020dual}. 

For the concerned MRC challenge with unanswerable questions, a reader has to handle two aspects carefully: 1) give the accurate answers for answerable questions; 2) effectively distinguish the unanswerable questions, and then refuse to answer. Such requirements lead to the recent reader's design by introducing an extra verifier module or answer-verification mechanism. Most readers simply stack the verifier along with encoder and decoder parts in a pipeline or concatenation way (Figure \ref{fig:overview}-[b-c]), which is shown suboptimal for installing a verifier.

As a natural practice of how humans solve complex reading comprehension \cite{DBLP:conf/sigir/ZhengMLYZM19,doi:10.1080/00461520.1987.9653053}, the first step is to read through the full passage along with the question and grasp the general idea; then, people re-read the full text and verify the answer if not so sure. Inspired by such a reading and comprehension pattern, we proposed a retrospective reader (Retro-Reader, Figure \ref{fig:overview}-[d]) that integrates two stages of reading and verification strategies: 1) sketchy reading that briefly touches the relationship of passage and question, and yields an initial judgment; 2) intensive reading that verifies the answer and gives the final prediction. Our major contributions are three folds:\footnote{Our source code is available at \url{https://github.com/cooelf/AwesomeMRC}.}

\begin{enumerate}
	\item We propose a new retrospective reader design which is capable of effectively performing answer verification instead of simply stacking verifier in existing readers.
	\item Experiments show that our reader can yield substantial improvements over strong baselines and achieve new state-of-the-art results on benchmark MRC tasks.
	\item For the first time, we apply the significance test for the concerned MRC task and show that our models are significantly better than the baselines.
\end{enumerate}

\section{Related Work}
The research of machine reading comprehension have attracted great interest with the release of a variety of benchmark datasets \cite{hill2015goldilocks,hermann2015teaching,Rajpurkar2016SQuAD,Joshi2017TriviaQA,Rajpurkar2018Know,lai2017race}. 
The early trend is a variety of attention-based interactions between passage and question, 
including Attention Sum \cite{kadlec2016text}, Gated attention \cite{Dhingra2017Gated}, Self-matching  \cite{Wang2017Gated}, Attention over Attention \cite{Cui2017Attention} and Bi-attention  \cite{Seo2016Bidirectional}.
Recently, PrLMs dominate the encoder design for MRC and achieve great success. These PrLMs include ELMo \cite{peters2018deep}, GPT \cite{radford2018improving}, BERT \cite{devlin2018bert}, XLNet \cite{yang2019xlnet},  RoBERTa \cite{liu2019roberta}, ALBERT \cite{Lan2020ALBERT}, and ELECTRA \cite{clark2019electra}. They show strong capacity for capturing the contextualized sentence-level language representations and greatly boost the benchmark performance of current MRC. 
Following this line, we take PrLMs as our backbone encoder.

In the meantime, the study of the decoder mechanisms has come to a bottleneck due to the already powerful PrLM encoder. 
Thus this work focuses on the non-encoder part, such as passage and question attention interactions, and especially the answer verification. 

To solve the MRC task with unanswerable questions is though important, only a few studies paid attention to this topic with straightforward solutions.
Mostly, a treatment is to adopt an extra answer verification layer, the answer span prediction and answer verification are trained jointly with multi-task learning (Figure \ref{fig:overview}-[c]). Such an implemented verification mechanism can also be as simple as an answerable threshold setting broadly used by powerful enough PrLMs for quickly building readers \cite{devlin2018bert,zhang2019semantics}.
\citeauthor{liu2018stochastic} \shortcite{liu2018stochastic} appended an empty word token to the context and added a simple classification layer to the reader. 
\citeauthor{hu2019read} \shortcite{hu2019read} used two types of auxiliary loss, independent span loss to predict plausible answers and independent no-answer loss the to decide answerability of the question. Further, an extra verifier is adopted to decide whether the predicted answer is entailed by the input snippets (Figure \ref{fig:overview}-[b]). 
\citeauthor{back2020neurquri} \shortcite{back2020neurquri} developed an attention-based satisfaction score to compare question embeddings with the candidate answer embeddings (Figure \ref{fig:overview}-[c]). 
\citeauthor{zhang2019sg} \shortcite{zhang2019sg} proposed a verifier layer, which is 
a linear layer applied to context embedding weighted by start and end distribution over the context words representations concatenated to \texttt{[CLS]} token representation for BERT (Figure \ref{fig:overview}-[c]).

Different from these existing studies which stack the verifier module in a simple way or just jointly learn answer location and non-answer losses, our Retro-Reader adopts a two-stage humanoid design \cite{DBLP:conf/sigir/ZhengMLYZM19,doi:10.1080/00461520.1987.9653053} based on a comprehensive survey over existing answer verification solutions.

\section{Our Proposed Model}
We focus on the span-based MRC task, which can be described as a triplet $\langle P, Q, A \rangle$, where $P$ is a passage, and $Q$ is a query over $P$, in which a span is a right answer $A$. Our system is supposed to not only predict the start and end positions in the passage $P$ and extract span as answer $A$ but also return a null string when the question is unanswerable. 

Our retrospective reader is composed of two parallel modules: a \textit{sketchy reading module} and an \textit{intensive reading module} to conduct a two-stage reading process. The intuition behind the design is that the sketchy reading makes a coarse judgment (external front verification) about the answerability, whether the question is answerable; and then the intensive reading jointly predicts the candidate answers and combines its answerability confidence (internal front verification) with the sketchy judgment score to yield the final answer (rear verification). 
\footnote{Intuitively, our model is supposed to be designed as shown in Figure \ref{fig:overview}-[d]. In the implementation, we find that modeling the entire reading process into two parallel modules is both simple and practicable with basically the same performance, which results in a parallel reading module design at last as shown in Figure \ref{fig:overview}-[e].}

\subsection{Sketchy Reading Module}
\paragraph{Embedding}
We concatenate question and passage texts as the input, which is firstly represented as embedding vectors to feed an encoder (i.e., a PrLM). In detail, 
the input texts are first tokenized to word pieces (subword tokens). Let $T=\{t_1,\dots,t_n\}$ denote a sequence of subword tokens of length $n$. 
For each token, the input embedding is the sum of its token embedding, position embedding, and token-type embedding. 
Let $X = \{x_1, \dots, x_n\}$ be the outputs of the encoder, which are embedding features of encoding sentence tokens of length $n$. The input embeddings are then fed to the interaction layer to obtain the contextual representations.

\paragraph{Interaction}
Following \citet{devlin2018bert}, the encoded sequence $X$ is processed to a multi-layer Transformer \cite{DBLP:conf/nips/VaswaniSPUJGKP17} for learning contextual representations. For the following part, we use $\textbf{H} = \{h_1, \dots, h_n\}$ to denote the last-layer hidden states of the input sequence.


\paragraph{External Front Verification}
After reading, the sketchy reader will make a preliminary judgment, whether the question is answerable given the context. We implement this reader as an external front verifier (E-FV) to identify unanswerable questions. The pooled first token (the special symbol, \texttt{[CLS]}) representation $h_1 \in \textbf{H}$, as the overall representation of the sequence, is passed to a fully connection layer to get classification logits $\hat{y}_{i}$ composed of answerable ($logit_{ans}$) and unanswerable ($logit_{na}$) elements. We use cross entropy as training objective: 
\begin{align}
\mathbb{L}^{ans} = -\frac{1}{N}\sum_{i=1}^{N}\left [y_{i}\log\hat{y}_{i} + (1-y_{i})\log(1-\hat{y}_{i} )\right ]
\end{align}
where $\hat{y}_{i} \propto \textup{SoftMax}(\textup{FFN}(h_1))$ denotes the prediction and $y_{i}$ is the target indicating whethter the question is answerbale or not. $N$ is the number of examples. We calculate the difference as the external verification score: $score_{ext} = logit_{na} - logit_{ans}$, which is used in the later rear verification as effective indication factor. 

\subsection{Intensive Reading Module}\label{sec:intensive}
The objective of the intensive reader is to verify the answerability, produce candidate answer spans, and then give the final answer prediction. It employs the same encoding and interaction procedure as the sketchy reader, to obtain the representation $\textbf{H}$. In previous studies \cite{devlin2018bert,yang2019xlnet,Lan2020ALBERT}, $\textbf{H}$ is directly fed to a linear layer to yield the prediction.

\paragraph{Question-aware Matching}
Inspired by previous success of explicit attention matching between passage and question \cite{kadlec2016text,Dhingra2017Gated,Wang2017Gated,Seo2016Bidirectional}, we are interested in whether the advance still holds based on the strong PrLMs. Here we investigate two alternative question-aware matching mechanisms as an extra layer. Note that this part is only used for ablation in Table \ref{tableatt} as a reference for interested readers. We do not use any extra matching part in our submission on test evaluations (e.g., in Tables \ref{tab:squad2.0}-\ref{tab:newsqa}) for the sake of simplicity as our major focus is the verification.

To obtain the representation of each passage and question, we split the last-layer hidden state $\textbf{H}$ into $\textbf{H}^Q$ and $\textbf{H}^P$ as the representations of the question and passage, according to its position information. Both of the sequences are padded to the maximum length in a minibatch.  
Then, we investigate two potential question-aware matching mechanisms, 1) Transformer-style multi-head \textit{cross attention} (CA) and 2) traditional \textit{matching attention} (MA).

$\bullet$ \textbf{Cross Attention} 
We feed the $\textbf{H}^Q$ and $\textbf{H}$ to a revised one-layer multi-head attention layer inspired by \citeauthor{lu2019vilbert} \shortcite{lu2019vilbert}. Since the setting is $\textbf{Q}=\textbf{K}=\textbf{V}$ in multi-head self attention,\footnote{In this work, $\textbf{Q},\textbf{K},\textbf{V}$ correspond to the items $Q_{m}^{l+1}x_{i}^{l}, K_{m}^{l+1}x_{j}^{l}$, $V_{m}^{l+1}x_{j}^{l}$, respectively.} which are all derived from the input sequence, we replace the input to $\textbf{Q}$ with $\textbf{H}$, and both of $\textbf{K}$ and $\textbf{V}$ with $\textbf{H}^Q$ to obtain the question-aware context representation $\textbf{H}'$.

$\bullet$ \textbf{Matching Attention}
Another alternative is to feed $\textbf{H}^Q$ and $\textbf{H}$ to a traditional matching attention layer \cite{Wang2017Gated}, by taking the question presentation $\textbf{H}^Q$ as the attention to the 
representation $\textbf{H}$:
\begin{equation}
	\begin{split}
	\textbf{M} &= \textup{SoftMax}(\textbf{H}(\textbf{W}\textbf{H}^Q+\textbf{b}\otimes\textbf{e})^\mathsf{T}),\\
	\textbf{H}' &= \textbf{M}\textbf{H}^Q,
	\end{split}
	\label{eq2:Image_Representation}
	\end{equation}
	where $\textbf{W}$ and $\textbf{b}$ are learnable parameters. $\textbf{e}$ is a all-ones vector and used to repeat the bias vector into the matrix. $\textbf{M}$ denotes the weights assigned to the different hidden states in the concerned two sequences. $\textbf{H}'$ is the weighted sum of all the hidden states and it represents how the vectors in $\textbf{H}$ can be aligned to each hidden state in $\textbf{H}^Q$. Finally, the representation $\textbf{H}'$ is used for the later predictions. If we do not use the above matching like the original use in BERT models, then $\textbf{H}' = \textbf{H}$ for the following part.

\paragraph{Span Prediction}
The aim of span-based MRC is to find a span in the passage as answer, thus we employ a linear layer with SoftMax operation and feed $\textbf{H}'$ as the input to obtain the start and end probabilities, $s$ and $e$:
\begin{equation}
s, e \propto \textup{SoftMax}(\textup{FFN}(\textbf{H}')).
\end{equation}

The training objective of answer span prediction is defined as cross entropy loss for the start and end predictions, 
\begin{equation}
\mathbb{L}^{span} = -\frac{1}{N}\sum_{i}^{N}[\log (p_{y^{s}_{i}}^s)+\log (p_{y^{e}_{i}}^e)]
\end{equation}
where $y^{s}_{i}$ and $y^{e}_{i}$ are respectively ground-truth start and end positions of example $i$. $N$ is the number of examples.

\paragraph{Internal Front Verification}\label{RV}
We adopted an internal front verifier (I-FV) such that the intensive reader can identify unanswerable questions as well. In general, a verifier's function can be implemented as  a cross-entropy loss (I-FV-CE), binary cross-entropy loss (I-FV-BE), or regression-style mean square error loss (I-FV-MSE). 
The pooled representation $h^{'}_{1} \in \textbf{H}'$, is passed to a fully connected layer to get the classification logits or regression score.  Let $\hat{y}_{i}  \propto \textit{Linear}(h_1)$ denote the prediction and $y_{i}$ is the answerability target, the three alternative loss functions are as defined as follows:

(1) We use cross entropy as loss function for the classification verification: 
\begin{equation}
\begin{split}
\bar{y}_{i,k} &= \textup{SoftMax}(\textup{FFN}(h^{'}_{1})),\\
\mathbb{L}^{ans} &= -\frac{1}{N}\sum_{i=1}^{N}\sum_{k=1}^{K}\left [ {y}_{i,k}\log\bar{y}_{i,k} \right ],
\end{split}
\end{equation}
where $K$ means the number of classes ($K=2$ in this work). $N$ is the number of examples.

(2) For binary cross entropy as loss function for the classification verification: 
\begin{equation}
\begin{split}
\bar{y}_{i} &= \textup{Sigmoid}(\textup{FFN}(h_1)),\\
\mathbb{L}^{ans} &= -\frac{1}{N}\sum_{i=1}^{N}\left [ y_{i}\log\bar{y}_{i} + (1-y _{i})\log(1-\bar{y}_{i})\right ].
\end{split}
\end{equation}

(3) For the regression verification, 
mean square error is adopted as its loss function: 
\begin{align}
\bar{y}_{i} &= \textup{FFN}(h^{'}_{1}),\\
\mathbb{L}^{ans}&= \frac{1}{N}\sum_{i=1}^{N}(y_{i}-\bar{y}_{i})^{2}.
\end{align}

During training, the joint loss function for FV is the weighted sum of the span loss and verification loss:
\begin{align}
\mathbb{L} = \alpha_{1}\mathbb{L}^{span} + \alpha_{2} \mathbb{L}^{ans},
\end{align}
where $\alpha_{1}$ and $\alpha_{2}$ are weights.

\paragraph{Threshold-based Answerable Verification}\label{sec:ap}
Following previous studies \cite{devlin2018bert,yang2019xlnet,liu2019roberta,Lan2020ALBERT}, we adopt threshold based answerable verification (TAV), which is a heuristic strategy to decide whether a question is answerable according to the predicted answer start and end logits finally. Given the output start and end probabilities $s$ and $e$, and the verification probability $v$, we calculate the has-answer score $score_{has}$ and the no-answer score $score_{null}$:
\begin{equation}
\begin{split}
score_{has} & =\max (s_k + e_l),1 < k \le l \le n, \\
score_{null} & = s_1+e_1,
\end{split}
\end{equation}

We obtain a difference score between $score_{null}$ and the $score_{has}$ as the final \textit{no-answer} score: $score_{diff} = score_{null} - score_{has}$. An answerable threshold $\delta$ 
is set and determined 
according to the development set. The model predicts the answer span that gives the \textit{has-answer} score if the final score is above the threshold $\delta$, and null string otherwise.

TAV is used in all our models as the last step for the answerability  decision. We denote it in our baselines with (+TAV) as default in Table \ref{tab:squad2.0}-\ref{tab:newsqa}, and omit the notation for simplicity in analysis part to avoid misunderstanding to keep on the specific ablations.

\subsection{Rear Verification}
Rear verification (RV) is the combination of predicted probabilities of E-FV and I-FV, which is an aggregated verification for final answer. 
\begin{equation}
v = \beta_{1} score_{diff} + \beta_{2} score_{ext},
\end{equation}
where $\beta_{1}$ and $\beta_{2}$ are weights. Our model predicts the answer span if $v > \delta$, and null string otherwise.

\section{Experiments}
\subsection{Setup}
We use 
the available PrLMs as encoder to build baseline MRC models: BERT \cite{devlin2018bert}, ALBERT \cite{Lan2020ALBERT}, and ELECTRA \cite{clark2019electra}. Our implementations of BERT and ALBERT are based on the public Pytorch implementation from Transformers.\footnote{\url{https://github.com/huggingface/transformers}.} ELECTRA is based on the Tensorflow release.\footnote{\url{https://github.com/google-research/electra}.} We use the pre-trained LM weights in the encoder module of our reader, using all the official hyperparameters.\footnote{BERT$_\texttt{large}$; ALBERT$_\texttt{xxlarge}$; ELECTRA$_\texttt{large}$.} For the fine-tuning in our tasks, we set the initial learning rate in \{2e-5, 3e-5\} with a warm-up rate of 0.1, and L2 weight decay of 0.01. The batch size is selected in \{32, 48\}. The maximum number of epochs is set in 2 for all the experiments. Texts are tokenized using wordpieces \cite{wu2016google}, with a maximum length of 512. Hyper-parameters were selected using the dev set. The manual weights are  $\alpha_{1}=\alpha_{2}=\beta_{1}=\beta_{2}=0.5$ in this work.


For answer verification, 
we follow the same setting according to the corresponding literatures \cite{devlin2018bert,Lan2020ALBERT,clark2019electra}, which simply adopts the answerable threshold method described in \S\ref{sec:ap}. In the following part, +TAV (for all the baseline modes) denotes the baseline verification for easy reference, which is equivalent to the baseline implementations in public literatures \cite{devlin2018bert,Lan2020ALBERT,clark2019electra}.

\subsection{Benchmark Datasets}
Our proposed reader is evaluated in two benchmark MRC challenges.
\paragraph{SQuAD2.0}
As a widely used MRC benchmark dataset, 
SQuAD2.0  \cite{Rajpurkar2018Know} combines the 100,000 questions in SQuAD1.1 \cite{Rajpurkar2016SQuAD} with over 50,000 new, unanswerable questions that are written adversarially by crowdworkers to look similar to answerable ones. 
The training dataset contains 87$k$ answerable and 43$k$ unanswerable questions.

\paragraph{NewsQA}
NewsQA 
\cite{trischler2017newsqa} 
is a question-answering dataset with 100,000 human-generated question-answer pairs. The questions and answers are based on a set of over 10,000 news articles from CNN supplied by crowdworkers. The paragraphs are about 600 words on average, which tend to be longer than SQuAD2.0. The training dataset has
20$k$ unanswerable questions among 97$k$ questions.

\subsection{Evaluation}

\paragraph{Metrics} Two official metrics are used to evaluate the model performance: Exact Match (EM) and a softer metric F1 score, which measures the average overlap between the prediction and ground truth answer at the token level.

\begin{table}
	\centering
	\setlength{\tabcolsep}{5pt}
	{
		\begin{tabular}{l c c c c }
			\toprule
			\multirow{2}{*}{\textbf{Model} }& \multicolumn{2}{c}{\textbf{Dev}} & \multicolumn{2}{c}{\textbf{Test}}\\
			& \textbf{EM} & \textbf{F1}&   \textbf{EM} & \textbf{F1}\\
			\midrule
			Human& -  & - & 86.8&    89.5 \\
			\midrule
			BERT \cite{devlin2018bert} & -  & - &  82.1 &  84.8 \\
			NeurQuRI \cite{back2020neurquri} & 80.0  & 83.1 & 81.3  & 84.3 \\
			XLNet \cite{yang2019xlnet}& 86.1  & 88.8  & 86.4 &    89.1   \\
			RoBERTa \cite{liu2019roberta} & 86.5  & 89.4 &  86.8 &  89.8 \\
			SG-Net \cite{zhang2019sg} & -  & - & 87.2 &    90.1 \\
			ALBERT \cite{Lan2020ALBERT}& 87.4  & 90.2 & 88.1 &    90.9 \\
			ELECTRA \cite{clark2019electra} & 88.0 & 90.6 & 88.7 & 91.4 \\
			\midrule
			\multicolumn{5}{c}{\emph{Our implementation}} \\
			ALBERT (+TAV) & 87.0 & 90.2 & - & -\\
			Retro-Reader on ALBERT & 87.8 & 90.9 & 88.1 & 91.4 \\
			ELECTRA (+TAV) & 88.0 & 90.6  & -  & - \\
			Retro-Reader on ELECTRA & \textbf{88.8} & \textbf{91.3} & \textbf{89.6} & \textbf{92.1} \\
			\bottomrule
		\end{tabular}
	}
	\caption{\label{tab:squad2.0} The results (\%) for SQuAD2.0 dataset. The results  are from the official leaderboard.  
	TAV: threshold-based answerable verification (\S\ref{sec:ap}). 
	}
\end{table}

\paragraph{Significance Test} With the rapid development of deep MRC models, the dominant models have achieved very high results (e.g., over 90\% F1 scores on SQuAD2.0), and further advance has been very marginal. 
Thus a significance test would be beneficial for measuring the difference in model performance. 

For selecting evaluation metrics for the significance test, since answers vary in length, using the F1 score would have a bias when comparing models, i.e., if one model fails on one severe example though works well on the others. Therefore, we use the tougher metric EM as the goodness measure. If the EM is equal to 1, the prediction is regarded as right and vice versa. Then the test is modeled as a binary classification problem to estimate the answer of the model is exactly right (EM=1) or wrong (EM=0) for each question. According to our task setting, we used McNemar’s test \cite{mcnemar1947note} to test the statistical significance of our results. This test is designed for paired nominal observations, and it is appropriate for binary classification tasks \cite{ziser2016neural}. 


The $p$-value is defined as the probability, under the null hypothesis, of obtaining a result equal to or more extreme than what was observed. The smaller the p-value, the higher the significance. A commonly used level of reliability of the result is 95\%, written as $p=0.05$.

\begin{table}
	\centering
	\setlength{\tabcolsep}{4.2pt}
	{
		\begin{tabular}{l c c c c }
			\toprule
			\multirow{2}{*}{\textbf{Model} }& \multicolumn{2}{c}{\textbf{Dev}} & \multicolumn{2}{c}{\textbf{Test}}\\
			& \textbf{EM} & \textbf{F1}&   \textbf{EM} & \textbf{F1}\\
			\midrule
			\multicolumn{5}{c}{\emph{Existing Systems}} \\
			BARB \cite{trischler2017newsqa} &36.1  & 49.6 & 34.1 & 48.2  \\
			mLSTM \cite{DBLP:conf/iclr/Wang017a} & 34.4 & 49.6 & 34.9 & 50.0 \\
			BiDAF \cite{Seo2016Bidirectional} & - & - & 37.1 & 52.3  \\
			R2-BiLSTM \cite{DBLP:journals/corr/Weissenborn17} & - & - & 43.7 & 56.7  \\
			AMANDA \cite{DBLP:conf/aaai/KunduN18} & 48.4& 63.3 &48.4 &63.7  \\
			DECAPROP \cite{tay2018densely} & 52.5 &65.7 &53.1 &66.3 \\
			BERT \cite{devlin2018bert} & - & - & 46.5 & 56.7 \\
			NeurQuRI \cite{back2020neurquri} & - & - & 48.2 & 59.5 \\
			\midrule
			\multicolumn{5}{c}{\emph{Our implementation}} \\
			ALBERT (+TAV) & 57.1 & 67.5 & 55.3 & 65.9\\
			Retro-Reader on ALBERT  &  \textbf{58.5} & \textbf{68.6} & \textbf{55.9} &  \textbf{66.8} \\
			ELECTRA (+TAV)  & 56.3 & 66.5 & 54.0 & 64.5\\
			Retro-Reader on ELECTRA & 56.9 &  67.0 & 54.7& 65.7 \\
			\bottomrule
		\end{tabular}
	}
	\caption{Results (\%) for NewsQA dataset. The results except ours are from \citeauthor{tay2018densely} \shortcite{tay2018densely} and \citeauthor{back2020neurquri} \shortcite{back2020neurquri}. TAV: threshold based answerable verification (\S\ref{sec:ap}).}\label{tab:newsqa} 
\end{table}

\subsection{Results}
Tables \ref{tab:squad2.0}-\ref{tab:newsqa} compare the leading single models on SQuAD2.0 and NewsQA.
Retro-Reader on ALBERT and Retro-Reader on ELECTRA denote our final models (i.e., our submissions to SQuAD2.0 online evaluation), which are respectively the ALBERT and ELECTRA based retrospective reader composed of both sketchy and intensive reading modules without question-aware matching for simplicity. 
According to the results, we make the following observations:

1) Our implemented ALBERT and ELECTRA baselines show the similar EM and F1 scores with the original numbers reported in the corresponding papers \cite{Lan2020ALBERT,clark2019electra}, ensuring that the proposed method can be fairly evaluated over the public strong baseline systems.

2) In terms of powerful enough PrLMs like ALBERT and ELECTRA, our Retro-Reader not only significantly outperforms the baselines with p-value $<0.01$,\footnote{Besides the McNemar's test, we also used paired t-test for significance test, with consistent findings.} but also achieves new state-of-the-art on the SQuAD2.0 challenge.\footnote{When our models were submitted (\textit{Jan 10th 2020} and \textit{Apr 05, 2020} for ALBERT- and ELECTRA-based models, respectively), our Retro-Reader achieved the first place on the SQuAD2.0 Leaderboard (https://rajpurkar.github.io/SQuAD-explorer/ ) for both single and ensemble models.}

3) The results on NewsQA further verifies the general
effectiveness of our proposed Retro-Reader. Our method shows consistent improvements over the baselines and achieves new state-of-the-art results.


\begin{table}
	\begin{center}
		\setlength{\tabcolsep}{5pt}
		{
			\begin{tabular}{lllllll}
				\toprule
				\multirow{2}{*}{\textbf{Model}} & \multicolumn{2}{c}{\textbf{All}} & \multicolumn{2}{c}{\textbf{HasAns}} & \multicolumn{2}{c}{\textbf{NoAns}} \\
				& \textbf{EM} & \textbf{F1} & \textbf{EM} & \textbf{F1} & \textbf{EM} & \textbf{F1} \\
				
				\midrule
				BERT  & 78.8 & 81.7  & 74.6 & 80.3 & 83.0 & 83.0 \\
				\quad     + E-FV  & 79.1 & 82.1 & 73.4 & 79.4  & 84.8 & 84.8  \\
				\quad     + I-FV-CE  & 78.6 & 82.0 & 73.3 & 79.5 & 84.5 & 84.5\\
				\quad + I-FV-BE & 78.8 & 81.8 & 72.6 & 78.7 & 85.0 & 85.0\\
				\quad     + I-FV-MSE  & 78.5 & 81.7& 73.0 & 78.6 & 84.8 & 84.8  \\
				\quad     + RV  & 79.6 & 82.5 & 73.7 & 79.6  & 85.2 & 85.2  \\
				\midrule 
				ALBERT  &  87.0 & 90.2 & 82.6 & 89.0 & 91.4 &  91.4  \\
				\quad     + E-FV  & 87.4 & 90.6  & 82.4 & 88.7  & 92.4 &  92.4\\
				\quad   + I-FV-CE  & 87.2 & 90.3 & 81.7 & 87.9 & 92.7 & 92.7 \\
				\quad + I-FV-BE & 87.2 & 90.2 &82.2 &88.4 &92.1 &92.1 \\
				\quad     + I-FV-MSE & 87.3 & 90.4 & 82.4 & 88.5 & 92.3 & 92.3 \\
				\quad      + RV  & \text{87.8} & \text{90.9}  & 83.1 & 89.4 &92.4  &  92.4 \\
				\bottomrule 
			\end{tabular}
		}
	\end{center}
	\caption{\label{tablescore} Results (\%) with different answer verification methods on the SQuAD2.0 dev set. \textit{CE}, \textit{BE}, and \textit{MSE} are short for the two classification and one regression loss functions defined in \S\ref{RV}.}
\end{table}

\begin{table}
	\begin{center}
		\setlength{\tabcolsep}{3.5pt}
		{
			\begin{tabular}{lllll}
				\toprule
				\textbf{Method}	& \textbf{Prec.} & \textbf{Rec.} 	& \textbf{F1} & \textbf{Acc.} \\
				\midrule
				ALBERT & 91.70 & 93.42 & 92.55 & 86.14  \\
				Retro-Reader on ALBERT & \textbf{94.30} & 92.38 & 93.33 & 87.49 \\
                ELECTRA & 92.71 & 92.58 & 92.64 & 86.30 \\
                Retro-Reader on ELECTRA & 93.27 & \textbf{93.51} & \textbf{93.39} & \textbf{87.60} \\
				\bottomrule
			\end{tabular}
		}
	\end{center}
	\caption{\label{tab:unans} Performance on the unanswerable questions from SQuAD2.0 dev set.}
\end{table}

\section{Ablations}

\subsection{Evaluation on Answer Verification}
Table \ref{tablescore} presents the results with different answer verification methods. We observe that either of the front verifiers boosts the baselines, and integrating both as rear verification works the best. Note that we show the HasAns and NoAns only for completeness. Since the final predictions are based on the threshold search of answerability scores (\S\ref{sec:ap}), there exists a tradeoff between the HasAns and NoAns accuracies. We see that the final RV that combines E-FV and I-FV shows the best performance, which we select as our final implementation for testing.

We further conduct the experiments on our model performance of the 5,945 unanswerable questions from the SQuAD 2.0 dev set. Results in Table \ref{tab:unans} show that our method improves the performance on unanswerable questions by a large margin, especially in the primary F1 and accuracy metrics.

\subsection{Comparisons with Equivalent Parameters}
When using sketchy reading module for external verification, we have two parallel modules that have independent parameters. For comparisons with equivalent parameters, we add an ensemble of two baseline models, to see if the advance is purely from the increase of parameters. Table \ref{tableequ} shows the results. We see that our model can still outperform two ensembled models. Although the two modules share the same design of the Transformer encoder, the training objectives (e.g., loss functions) are quite different, one for answer span prediction, the other for answerable decision. The results indicate that our two-stage reading modules would be more effective for learning diverse aspects (verification and span prediction) for solving MRC tasks with different training objectives. From the two modules, we can easily find the effectiveness of either the span prediction or answer verification, to improve the modules correspondingly. We believe this design would be quite useful for real-world applications.

\begin{table}
	\begin{center}
		\setlength{\tabcolsep}{13.5pt}
		{
			\begin{tabular}{lll}
				\toprule
				\textbf{Method}	& \textbf{EM} & \textbf{F1} \\
				\midrule
				ALBERT & 87.0 & 90.2  \\
				Two-model Ensemble & 87.6  &  90.6    \\
				Retro-Reader  &  87.8 & 90.9  \\
				\bottomrule
			\end{tabular}
		}
	\end{center}
	\caption{\label{tableequ} Comparisons with Equivalent Parameters on the dev set of SQuAD2.0.}
\end{table}

\begin{table}
	\begin{center}
		\setlength{\tabcolsep}{10pt}
		{
			\begin{tabular}{lllll}
				\toprule
				\multirow{2}{*}{\textbf{Method}} & \multicolumn{2}{c}{\textbf{SQuAD2.0}} & \multicolumn{2}{c}{\textbf{NewsQA}} \\
				& \textbf{EM} & \textbf{F1} & \textbf{EM} & \textbf{F1} \\
				\midrule
				BERT &  78.8 & 81.7 & 51.8 & 62.5 \\
				\quad     + CA &  78.8 & 81.7 & 52.1 & 62.7\\
				\quad     + MA  &  78.3  & 81.4  & 52.4 & 62.6 \\
				\midrule 
				ALBERT & 87.0 & 90.2 & 57.1 & 67.5\\
				\quad     + CA & 87.3 & 90.3  & 56.0 & 66.3  \\
				\quad     + MA  & 86.8 & 90.0  & 55.8 & 66.1  \\
				\bottomrule
			\end{tabular}
		}
	\end{center}
	\caption{\label{tableatt} Results (\%) with matching interaction methods on the dev sets of SQuAD2.0 and NewsQA.}
\end{table}
\subsection{Evaluation on Matching Interactions}
Table \ref{tableatt} shows the results with different interaction methods described in \S\ref{sec:intensive}. 
We see that merely adding extra layers could not bring noticeable improvement, which indicates that simply adding more layers and parameters would not substantially benefit the model performance. The results verified the PrLMs' strong ability to capture the relationships between passage and question after processing the paired input by deep self-attention layers. In contrast, answer verification could still give consistent and substantial advance.

\begin{table}
	\begin{center}
		\begin{tabular}{|p{8cm}|}
			\hline
			\textbf{Passage}: \\
			\textit{Southern California consists of a heavily developed urban environment, home to some of the largest urban areas in the state, along with vast areas that have been left undeveloped. \textcolor[RGB]{0,145,147}{It is the third most populated megalopolis in the United States, after the Great Lakes Megalopolis and the Northeastern megalopolis.} Much of southern California is famous for its large, spread-out, suburban communities and use of automobiles and highways. The dominant areas are Los Angeles, Orange County, San Diego, and Riverside-San Bernardino, each of which are the centers of their respective metropolitan areas...}\\ 
			\hline
			\textbf{Question}: \\
			\textit{What are the second and third most populated megalopolis after Southern California}?\\    
			\hline
			\textbf{Answer:} \\
			\textbf{Gold: } $\langle\textup{no answer}\rangle$ \\
			\textbf{ALBERT (+TAV)}: Great Lakes Megalopolis and the Northeastern megalopolis. \\
			\textbf{Retro-Reader over ALBERT:} $\langle\textup{no answer}\rangle$ \\
			$score_{has}=0.03, score_{na}=1.73, \delta=-0.98$ \\
			\hline
		\end{tabular}
	\end{center}
	\caption{\label{com_exps} Answer prediction examples from the ALBERT baseline and Retro-Reader.}
\end{table}

\subsection{Comparison of Predictions}
To have an intuitive observation of the predictions of Retro-Reader, we give a prediction example on SQuAD2.0 from baseline and Retro-Reader in Table \ref{com_exps}, which shows that our method works better at judging whether the question is answerable on a given passage and gets rid of the plausible answer.

\section{Conclusion}
As machine reading comprehension tasks with unanswerable questions stress the importance of answer verification in MRC modeling, this paper devotes itself to better verifier-oriented MRC task-specific design and implementation for the first time. Inspired by human reading comprehension experience, we proposed a retrospective reader that integrates both sketchy and intensive reading. With the latest PrLM as encoder backbone and baseline, the proposed reader is evaluated on two benchmark MRC challenge datasets SQuAD2.0 and NewsQA, achieving new state-of-the-art results and outperforming strong baseline models in terms of newly introduced statistical significance, which shows the choice of verification mechanisms has a significant impact for MRC performance and verifier is an indispensable reader component even for powerful enough PrLMs used as the encoder. In the future, we will investigate more decoder-side problem-solving techniques to cooperate with the strong encoders for more advanced MRC.

\bibliography{retro}

\end{document}